\documentclass{article}

\PassOptionsToPackage{numbers, compress}{natbib}
\usepackage[preprint]{neurips_2026}

\usepackage[utf8]{inputenc}
\usepackage[T1]{fontenc}   
\usepackage{hyperref}      
\usepackage{url}            
\usepackage{booktabs}      
\usepackage{amsfonts}      
\usepackage{nicefrac}      
\usepackage{microtype}     
\usepackage{xcolor}        
\usepackage{amsmath}
\usepackage{graphicx}
\usepackage{caption}
\usepackage{wrapfig}
\usepackage{pifont}
\usepackage{makecell}
\usepackage{subfigure}
\usepackage{titletoc}
\usepackage{minitoc}

\title{From Zero to Hero: Training-Free Custom Concept Spawning in World Models}

\author{
  Kiymet Akdemir \quad \quad  Pinar Yanardag \\
  Virginia Tech \\
  \texttt{\href{https://spawn-world.github.io/}{spawn-world.github.io}}
}

\begin{document}

\maketitle

\begin{center}
    \centering
    \captionsetup{type=figure}
    \includegraphics[width=0.95\linewidth]{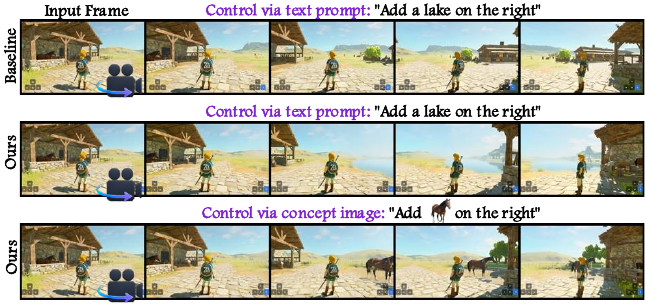}
    \captionof{figure}{\textbf{Controllable concept spawning with SPAWN.} Given an input 
frame and rightward camera motion, the baseline WorldPlay populates the unseen region with its priors, ignoring the user's request (top row). SPAWN instead renders the requested concept into the world via either a text prompt (middle row) or a concept image (bottom row), integrated with consistent lighting and 
perspective.}
    \label{fig:teaser}
\end{center}

\begin{abstract}

Autoregressive world models have emerged as a powerful paradigm for interactive video generation, allowing users to navigate dynamically generated environments through actions. These models are typically conditioned on a text prompt and/or a single reference frame, from which the entire world is generated. Yet the moment the user navigates beyond what is visible in that frame, the unseen regions are populated by the base model's priors, with no mechanism for the user to specify what should appear and where. This is a fundamental limitation for applications such as gaming, interactive storytelling, and simulation, where controllable scene composition is essential. We refer to this missing capability as concept spawning; introducing a user-specified visual concept into a world model, analogous to spawning in a game engine. We introduce SPAWN (\textbf{S}wapping \textbf{P}inned \textbf{A}nchor with \textbf{W}indowed i\textbf{N}jection), a training-free method for concept spawning. SPAWN exploits a structural property of image-to-video backbones: the first slot of the context memory is pinned to the reference frame and acts as a foundational anchor for every generated chunk. By swapping this anchor with an external concept latent over a short injection window and letting the original anchor return, we cause the concept to propagate naturally through the rollout via the model's own memory. SPAWN supports concepts from fine-grained entities such as characters and props to large-scale elements such as buildings and landmarks, and accepts either a concept image or a text description as input. Experiments show that SPAWN integrates concepts with consistent lighting, scale, and perspective while preserving identity and temporal coherence, demonstrating that controllable concept spawning is achievable in existing autoregressive world models without any training.
\end{abstract}
    
\section{Introduction}
\label{sec:intro}

Autoregressive world models have rapidly gained popularity for their potential to power immersive applications such as gaming, interactive storytelling, simulation, and embodied AI, where users can navigate dynamically generated environments through actions in real time ~\cite{bruce2024genie, che2024gamegen, sun2025worldplay, xiao2025worldmem}. These models are typically conditioned on a single reference frame or a text prompt, from which the entire surrounding world is generated chunk by chunk as the user navigates. The visible portion of the input frame anchors the scene in front of the camera. However, once the user turns to look elsewhere, the unseen regions must be hallucinated based on the base model’s priors. As a result, the model fills in the world with whatever it deems a plausible continuation of the scene, without any mechanism for the user to control what appears or where.  We refer to this missing capability as concept spawning: the ability to introduce a user-specified visual concept into an already-running world at a chosen moment, analogous to spawning in a game engine.

\setlength{\intextsep}{0pt}
\setlength{\columnsep}{8pt}
\begin{wraptable}{r}{0.55\linewidth}
\setlength{\tabcolsep}{2pt}
\small
\centering
\begin{tabular}{lccccc}
\toprule
Method & \makecell{Concept\\Spawning} & \makecell{Out-of-\\View} & \makecell{Image-\\Cond.} & \makecell{Training-\\Free} & \makecell{Public}\\
\midrule
WorldPlay & \ding{55} & \ding{55} & \ding{55} & \ding{55} & \checkmark\\
GameGen-X & \checkmark & \ding{55} & \ding{55} & \ding{55} & \ding{55}\\
GameCraft-2 & \checkmark & \ding{55} & \ding{55} & \ding{55} & \ding{55}\\
\textbf{SPAWN} & \checkmark & \checkmark & \checkmark & \checkmark & \checkmark\\
\bottomrule
\end{tabular}
\caption{Capability comparison across interactive world models. SPAWN is the only method that supports 
image-conditioned concept spawning without training.}
\label{tab:compare_methods}
\end{wraptable}

This is a fundamental limitation since the ability to compose and control scenes is central to level design, interactive storytelling, embodied AI simulation, and any application where generated worlds must be deliberately controlled instead of passively sampled.   Existing work has made progress toward controllability through mechanisms such as text-conditioned scene generation \cite{che2024gamegen} and instruction-following world models \cite{tang2025hunyuan}. However, these approaches provide only shallow control (See Tab.~\ref{tab:compare_methods}): they can instantiate concepts via text-prompts within the current field of view, but fail to maintain consistent, spatially grounded control as the user navigates the environment. Moreover, these methods do not support user-specified visual concepts, instead relying on abstract textual descriptions, and often require costly training on curated interactive data. 

We introduce SPAWN (\textbf{S}wapping \textbf{P}inned \textbf{A}nchor with \textbf{W}indowed i\textbf{N}jection), a training-free method for concept spawning in a world model. Our approach is motivated by a key observation: autoregressive world 
models built on image-to-video backbones have the first slot of the 
context memory pinned to the reference frame across the entire rollout, 
where it acts as a foundational anchor that every generated chunk 
attends to. This slot effectively acts as the model’s notion of ground truth for the scene, analogous to attention sinks observed in large language models, where specific tokens disproportionately anchor attention and shape downstream generation \cite{peng2026attention}. SPAWN exploits this property by swapping the pinned anchor with an external concept latent over a short windowed injection, then letting the original anchor return. During the injection window, the concept propagates into the temporal and geometric memory caches via the model's normal generation process; once the window closes and the original anchor is restored, the concept persists in those downstream caches and continues to appear naturally as the camera navigates the world. SPAWN does not require  architectural modifications,  fine-tuning, or curated training data.

A user can drive SPAWN with either a concept image, when they have a specific visual reference in mind (e.g., this specific horse), or a text description (e.g., a lake on the right).  Figure~\ref{fig:teaser} illustrates the resulting capability: given the same input frame and rightward camera motion, the baseline world model fills the unseen region with priors-driven content, while SPAWN renders the requested lake or specific horse directly into the scene, integrated with consistent lighting, scale, and perspective. We evaluate SPAWN on WorldPlay across 200 generated rollouts spanning diverse scenes, concept categories, and visual styles. SPAWN supports concepts from fine-grained entities such as characters and props to large-scale scene elements such as buildings and landscapes, and its effects persist across multi-second rollouts and multiple sequential injections. Our experiments show that SPAWN preserves visual identity, temporal coherence, and scene integration without any training or architectural changes to the base model, suggesting that controllable concept spawning is an inference-time capability of existing autoregressive world models. Our main contributions are as follows:

\begin{itemize}

 \item  We introduce SPAWN, a training-free mechanism that exploits the structural anchor role of the first context-memory slot to spawn concepts into a world with a user-provided concept image or a text-prompt.
\item We demonstrate on WorldPlay that SPAWN spawns concepts at arbitrary granularity with consistent lighting, scale, and perspective, preserves identity and temporal coherence across multi-second rollouts, and supports multiple sequential injections; all without any architectural modification or additional training.
\item We release our code and a benchmark dataset of 200 rollouts 
spanning diverse scenes and concept categories, enabling systematic 
evaluation of concept spawning in generated worlds.
\end{itemize}

\section{Related Work}
\noindent \textbf{Video Generation} Diffusion models \cite{ho2020denoising, song2020score, peebles2023scalable} 
are the dominant paradigm for visual generative modeling, with early video 
work adopting latent diffusion frameworks \cite{rombach2022high} for 
efficient, temporally coherent synthesis in compressed latent spaces 
\cite{blattmann2023stable, ho2022video, guo2023animatediff}. The field has 
since transitioned from U-Net to scalable Diffusion Transformer (DiT) 
backbones \cite{peebles2023scalable}, with large-scale models such as 
CogVideoX \cite{yang2024cogvideox}, Wan \cite{wan2025wan}, and 
HunyuanVideo \cite{kong2024hunyuanvideo} generating high-quality but 
non-interactive sequences. Autoregressive video generation 
\cite{yan2021videogpt, kim2024fifo, henschel2025streamingt2v} extends 
video length by streaming frame-by-frame; Diffusion Forcing 
\cite{chen2024diffusion} introduces per-frame noise levels in place of 
the uniform-noise full-sequence paradigm, and Self-Forcing 
\cite{huang2025self} closes the training-inference gap by conditioning 
on the model's own previously generated frames to reduce long-horizon 
error accumulation. Text-guided video editing 
\cite{qi2023fatezero, ceylan2023pix2video, liu2024video} similarly 
steers diffusion priors toward targeted scene modifications, though 
it operates on fixed sequences rather than interactive environments.

\noindent \textbf{Interactive World Models} World video models extend video generation to interactive settings, 
generating temporally coherent sequences conditioned on user actions. 
Early action-conditioned generators established the paradigm in narrow 
game domains with limited action vocabularies and short horizons 
\cite{yu2025gamefactory, mao2025yume}. Subsequent work expanded the 
interaction and environmental scope through multi-modal control 
\cite{che2024gamegen} and richer open-world environments 
\cite{zhang2025matrix, he2025matrix}. A parallel line addresses the long-horizon consistency that 
sliding-window approaches fail to preserve, through external memory 
banks \cite{xiao2025worldmem}, explicit 3D surface reconstruction 
\cite{li2025vmem}, retrieval-based context extension 
\cite{yu2025context}, and selective KV-cache management for prompt 
adaptability \cite{yang2025longlive}. WorldPlay 
\cite{sun2025worldplay}, retrieves 
geometrically relevant frames via camera pose overlap and 
reassigns their temporal embeddings to maintain consistency across 
extended rollouts. Most recently, the Hunyuan-GameCraft line 
\cite{li2025hunyuan, tang2025hunyuan} unifies discrete control with 
language-guided world steering, distinct 
from the autoregressive video generation we target. Despite this progress, no existing approach supports 
image-based concept customization. We close this gap with a training-free mechanism 
that replaces fixed anchor slots of the context memory with an external concept during autoregressive rollout.

\section{Method}
\label{sec:method}
\begin{figure}
    \centering
    \includegraphics[width=\linewidth]{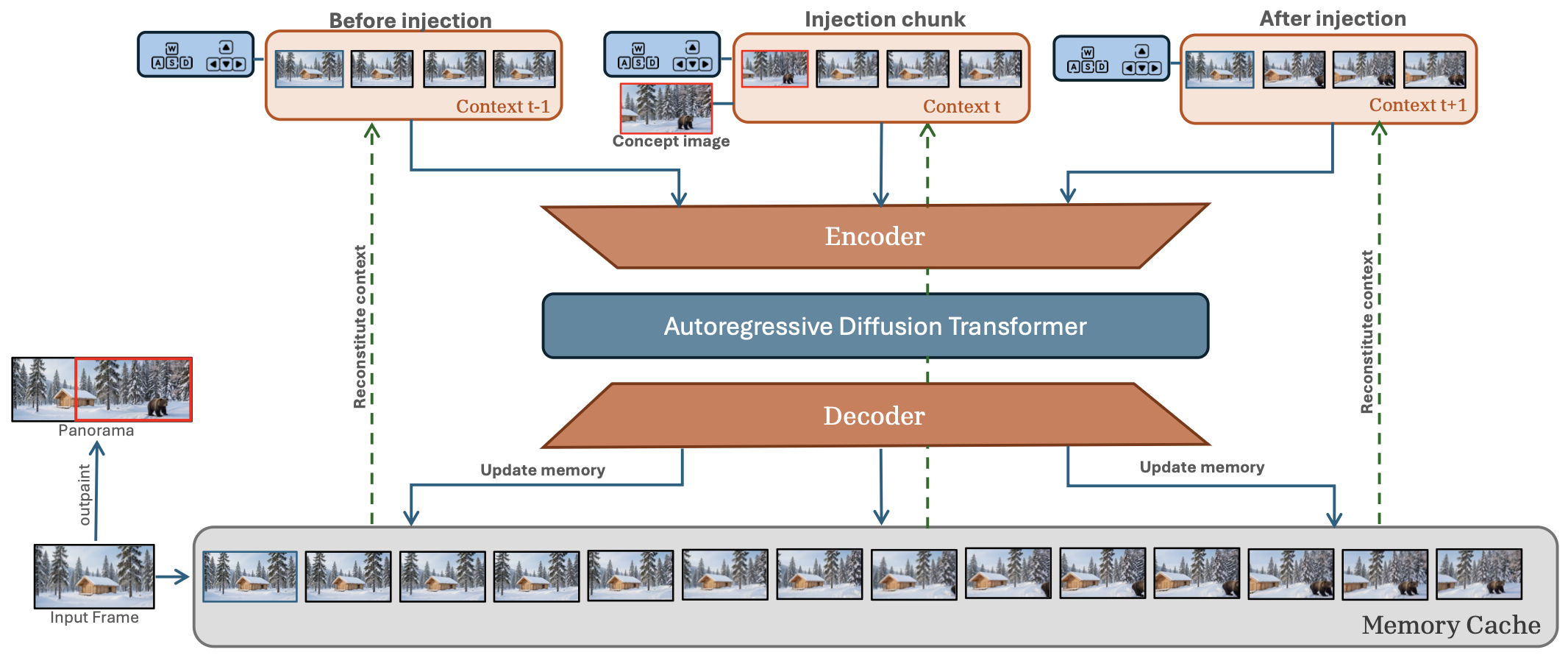}
    \caption{\textbf{Method overview.} At each rollout step, the context memory is reconstituted from the 
memory cache and passed through the encoder, autoregressive diffusion 
transformer, and decoder to produce the next chunk. At specified 
injection chunks, we replace the anchor slot of the context memory 
with a concept latent before encoding, which introduces the concept 
into the scene. Once the injection window ends, we restore the 
original anchor; even so, the concept persists in subsequent chunks. 
For clarity, the figure shows a reduced context size and a single 
replaced slot.}
    \label{fig:method}
    \vspace{-1em}
\end{figure}

 Our method spawns concepts by replacing the anchor slot of the context memory with a concept latent over a short injection window. We first describe the context-memory structure, then detail the injection mechanism. See Fig.~\ref{fig:method} for a method overview.

\subsection{Preliminaries}
\label{sec:preliminaries}

We build on WorldPlay~\cite{sun2025worldplay}, an autoregressive world model that performs next-chunk prediction $N_\theta(x_t \mid O_{t-1}, A_{t-1}, a_t, r)$, where $x_t \in \mathbb{R}^{C \times 4 \times H \times W}$ is the next chunk of four latents (16 frames after VAE decoding), $O_{t-1} = \{x_0, \ldots, x_{t-1}\}$ is the past, $A_{t-1} = \{a_0, \ldots, a_{t-1}\}$ and $a_t$ are past and current actions, and $r$ is a single reference image or text description of the world. The full rollout consists of $N$ chunks. When $r$ is an image, it is encoded by a causal 3D VAE $\mathcal{E}$ into the initial chunk $x_0$, which serves as the foundational observation seeding the rollout.

Conditioning on all of $O_{t-1}$ is intractable, so the model conditions on a reconstituted context memory
\begin{equation}
C_t = C_t^T \cup C_t^S, \qquad C_t^S \subseteq O_{t-1} \setminus C_t^T,
\end{equation}
where $C_t^T$ holds the most recent chunks for short-term motion smoothness and $C_t^S$ samples non-adjacent frames by camera-pose overlap and proximity for long-range geometric consistency. Frames in $C_t$ are ordered by their absolute temporal index, so the first slot is the oldest frame in context. Because the spatial-retrieval set is reselected at every chunk as the camera moves, the visual KV cache is recomputed from $C_t$ at the start of every chunk's denoising,
\begin{equation}
\mathrm{KV}_t \;\leftarrow\; N_\theta^{\mathrm{KV}}(C_t),
\label{eq:kv_recompute}
\end{equation}
where $N_\theta^{\mathrm{KV}}$ denotes the encoding pass that produces keys and values from the context latents.

\subsection{Swapping Pinned Anchor with Windowed Injection}
\label{sec:spawning}

\noindent\textbf{The anchor role of the first slot.}
In autoregressive video diffusion models with image conditioning, the first slot of the context memory is structurally distinguished. It holds the reference image as a fixed, zero-noise latent throughout the rollout~\cite{kong2024hunyuanvideo,yang2024cogvideox,wan2025wan}, serving as a fixed global anchor that every chunk attends to from the very first prediction to the last. Its content conditions every chunk the model generates, and its influence does not decay with rollout length; a behavior consistent with the attention-sink phenomenon originally observed in LLMs~\cite{xiao2023streamingllm}, where a small number of early positions disproportionately concentrate attention mass. Recent autoregressive video generation methods~\cite{yang2025longlive,liu2025rolling,yi2025deep} leverage the persistent influence of the first slots as a \emph{stabilizer}: a fixed early region of the KV cache is held constant throughout the rollout to prevent long-range collapse and preserve global consistency. We instead repurpose them as a control surface. By overwriting the latents at the anchor positions with an external concept before the chunk's KV recompute, that concept inherits the anchor role and propagates through the model's downstream generation as if it had been part of the reference scene from the start. Replacing any other slots alone does not produce this effect (See~\ref{supp:ablation}).

\begin{figure}
    \centering
    \includegraphics[width=\linewidth]{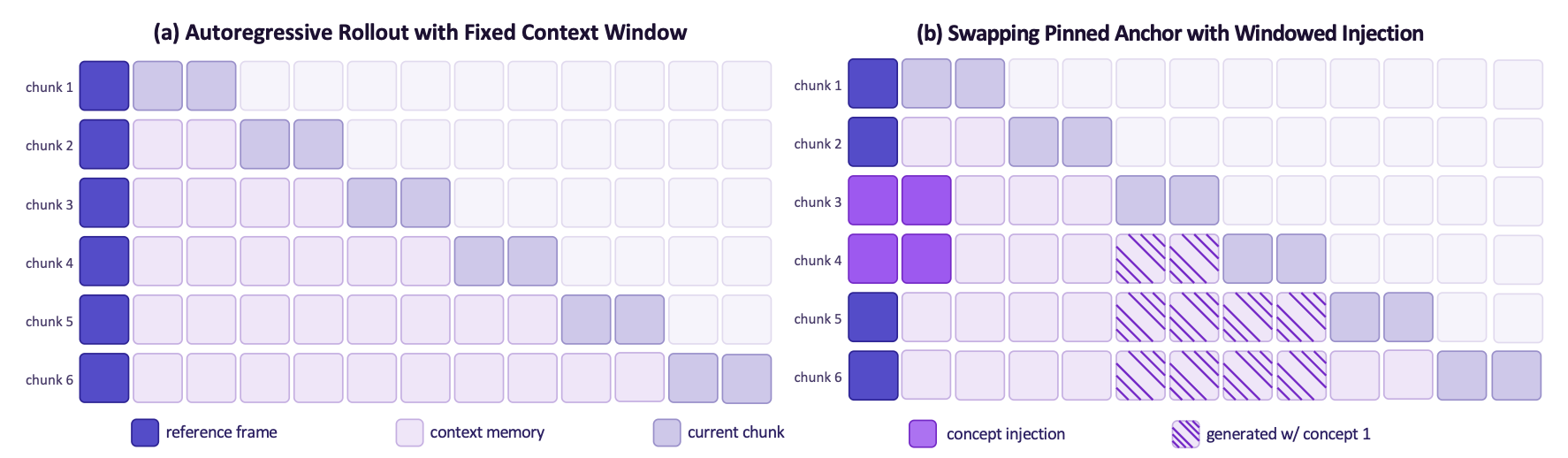}
    \caption{\textbf{(a) Standard rollout:} context memory evolves 
without intervention; the first slot holds the reference frame and 
anchors the rollout. \textbf{(b) SPAWN:} anchor slots are replaced with a 
concept latent during injection chunks, and diagonal hatching marks 
slots showing how the concept propagates through subsequent 
generation. The figure shows simplified memory size and concept 
counts.}
    \label{fig:grid}
    \vspace{-1em}
\end{figure}

\noindent\textbf{Anchor replacement.}
Let $I_c$ be a concept image and $z_c = \mathcal{E}(I_c) \in \mathbb{R}^{C \times 1 \times H \times W}$ its single-frame VAE encoding, in the same latent space as the chunks in $C_t$. Given a set of injection chunks $\mathcal{T} \subseteq \{1, \ldots, N\}$, at every $t \in \mathcal{T}$ we replace the anchor slot of $C_t$ in place with $z_c$:
\begin{equation}
C'_t = \bigl(z_c,\; C_t[1],\; \ldots,\; C_t[|C_t| - 1]\bigr).
\label{eq:anchor_replacement}
\end{equation}
In our experiments, we observe that this single substitution is 
sufficient to spawn the concept into the rollout. However, we extend 
the substitution to the second anchor slot, since we find that doing 
so strengthens identity preservation. All other slots, including the 
rest of the temporal memory and the geometrically retrieved spatial 
memory, are left unchanged, so scene context, motion continuity, and 
long-range geometric consistency are preserved.

\noindent\textbf{Recaching.}
With $C'_t$ in place of $C_t$, the per-chunk KV reset of Eq.~\ref{eq:kv_recompute} gives:
\begin{equation}
\mathrm{KV}_t \;\leftarrow\; N_\theta^{\mathrm{KV}}(C'_t).
\end{equation}
Each DiT block of $N_\theta$ computes self-attention with two complementary positional encodings, namely 3D RoPE for video tokens and PRoPE~\cite{li2025cameras} for camera frustums:
\begin{align}
\mathrm{Attn}_1 &= \mathrm{Attn}\bigl(R^\top \odot Q,\; R^{-1} \odot K,\; V\bigr), \\
\mathrm{Attn}_2 &= D_{\mathrm{proj}} \odot \mathrm{Attn}\bigl((D_{\mathrm{proj}})^\top \odot Q,\; (D_{\mathrm{proj}})^{-1} \odot K,\; (D_{\mathrm{proj}})^{-1} \odot V\bigr),
\end{align}
with block output $\mathrm{Attn}_1 + \mathrm{zero\_init}(\mathrm{Attn}_2)$. After the recompute over $C'_t$, the cached keys and values $K, V$ at the substituted anchor positions are computed by $N_\theta^{\mathrm{KV}}$ with $z_c$ as the anchor input. These are then RoPE-rotated under $R$ in $\mathrm{Attn}_1$ and frustum-projected under $D_{\mathrm{proj}}$ in $\mathrm{Attn}_2$ at attention time. The current chunk's queries $Q$ thus attend to a cache whose anchor region holds the concept, while every other position is identical to what the base model would have computed. The substitution requires no architectural change, no additional masking, and no fine-tuning, since $N_\theta$ already consumes an arbitrary $C_t$ at every chunk under both attention paths.

\noindent\textbf{Concept persistence after the window.}
Once the rollout passes the last chunk in $\mathcal{T}$, no further substitution takes place. For $t \notin \mathcal{T}$, the anchor slots of $C_t$ are repopulated by the standard reconstitution process, yet the concept persists in the rollout, for two reasons. First, the chunks generated during the window are themselves rendered with the concept present in the scene, and they enter $C_t^T$ for subsequent chunks through the standard temporal-memory pathway, carrying the concept forward as long as those chunks remain within the temporal window. Second, once those chunks fall out of $C_t^T$, the spatial-retrieval pathway $C_t^S$ surfaces them whenever the current camera returns to a viewpoint that overlaps with the chunks generated during injection. The concept therefore continues to appear naturally through the same memory mechanism that maintains long-term geometric consistency in the base model, with no explicit removal step or additional bookkeeping. In effect, the injection window seeds the model's memory; the rollout then propagates the concept on its own (Fig.~\ref{fig:grid}).

\noindent\textbf{Multi-concept spawning.}
The formulation extends to multiple concepts $\{(I_c^j, \mathcal{T}^j)\}_{j=1}^M$ with disjoint schedules $\mathcal{T}^j \cap \mathcal{T}^l = \emptyset$ for $j \neq l$. At each $t \in \mathcal{T}^j$, the anchor slot of $C_t$ is replaced with $z_c^j = \mathcal{E}(I_c^j)$ via Eq.~\ref{eq:anchor_replacement}. Concepts intended to appear together at the same chunk are combined into a single concept image and injected jointly. 

\noindent\textbf{Concept image.}
Our method allows the concept image to be constructed flexibly, 
either by outpainting~\cite{wu2025qwenimagetechnicalreport} the 
reference frame in the target viewing direction or by generating it 
via ControlNet~\cite{zhang2023adding}, conditioned on a text 
description of the concept or an image reference. We then crop the 
region corresponding to the spawn direction to obtain $I_c$. This serves two purposes: the placement of the concept in the world 
is precisely controlled by the user through the outpainting step, 
and the concept image is generated within the context of the 
surrounding scene.

\section{Experiments}

\begin{figure}
    \centering
    \includegraphics[width=\linewidth]{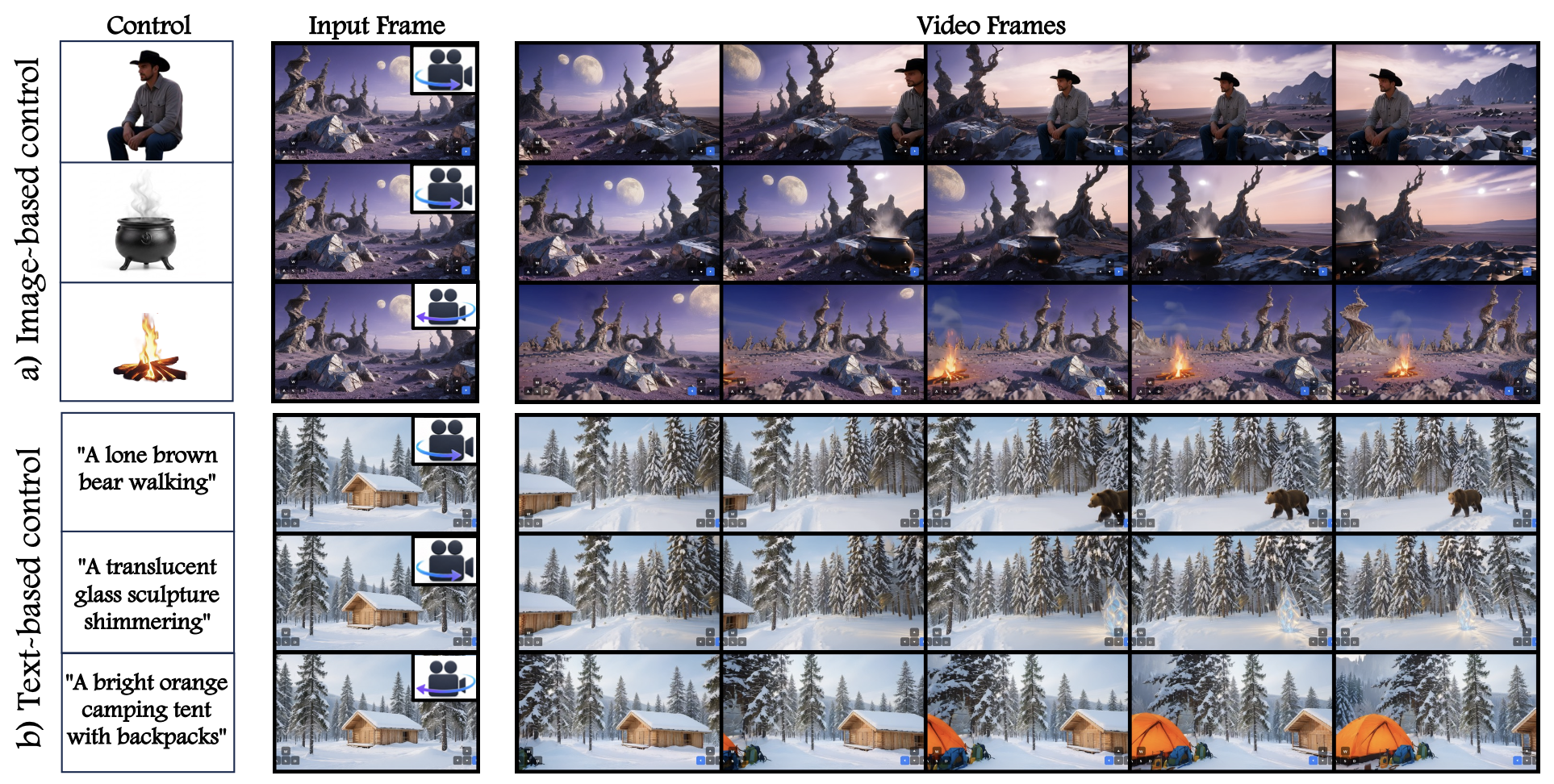}
    \caption{SPAWN supports both image-based (a) and text-based (b) control. Each 
row injects a concept, specified by an image or a text prompt, with consistent lighting and 
perspective.}
    \label{fig:qual_ours}
\end{figure}

\noindent\textbf{Baselines.}~ Several recent works share aspects of our 
setting but none target the exact same task: 
Hunyuan-GameCraft-2~\cite{tang2025hunyuan} introduces entities via training, 
GameGen-X~\cite{che2024gamegen} follows text-based instructions at inference. None are publicly available at the time of writing, ruling out direct comparison. We therefore evaluate against 
three available baselines: HunyuanVideo~\cite{kong2024hunyuanvideo} and 
Wan~2.2~\cite{wan2025wan}, two strong image-to-video (I2V) foundation models testing whether a general video backbone can introduce the requested concept 
under text conditioning alone, and WorldPlay~\cite{sun2025worldplay}, the autoregressive world model on which we build. All baselines receive a text prompt describing the scene with the concept 
present; for HunyuanVideo and Wan~2.2, which do not expose a separate action interface, the target camera motion is encoded into the same prompt.

\noindent\textbf{Evaluation set.}~ We build on WorldPlay~\cite{sun2025worldplay}, 
an open-source autoregressive world model based on the 
Hunyuan-Video~\cite{kong2024hunyuanvideo} backbone, using the publicly 
released checkpoint without modification or fine-tuning; all experiments 
run on two NVIDIA H200 GPUs. We evaluate on 200 generated rollouts of 
8 seconds each, spanning a range of scenes, concepts, and visual styles. 
The 8-second horizon is chosen for fair comparison with the I2V baselines 
below, which are trained for short-clip generation and do not support 
extended autoregressive rollout. See Appendix~\ref{supp:exp} for details.

\begin{figure}
    \centering
    \includegraphics[width=0.9\linewidth]{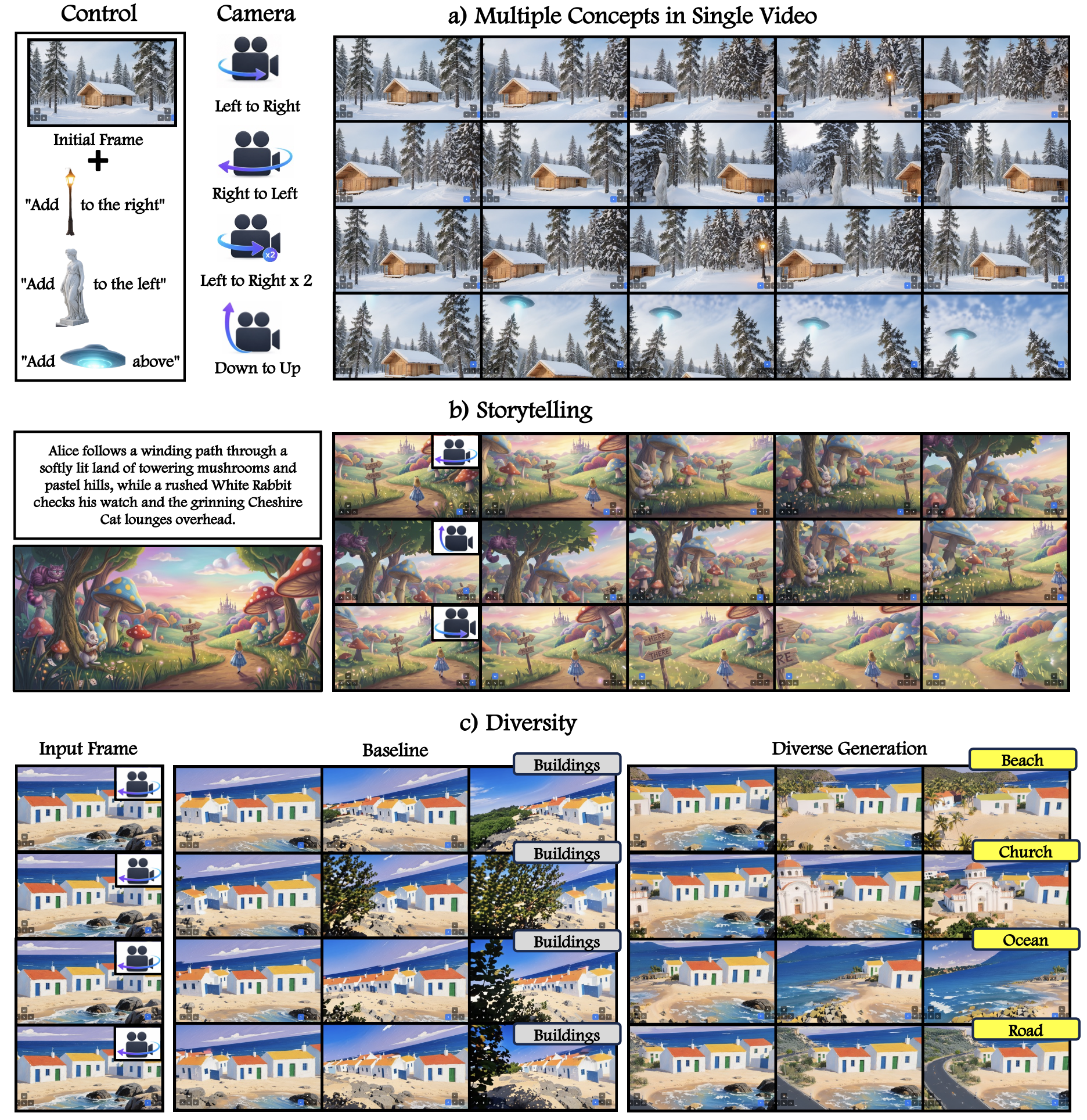}
    \caption{Use cases of SPAWN. \textbf{Multiple Concepts:} distinct concepts 
spawned into a single rollout. \textbf{Storytelling:} a story-derived 
panorama seeds both the reference image and the concepts that 
populate the world. \textbf{Diversity:} spawning populates the world 
with diverse landscape features.}
    \label{fig:use_case}
    \vspace{-0.5cm}
\end{figure}

\noindent\textbf{Metrics.}~ We report VBench~\cite{huang2023vbench, huang2025vbench++} 
dimensions covering general video quality (Aesthetic Quality, Imaging 
Quality), temporal coherence (Motion Smoothness, Subject Consistency, 
Background Consistency), and motion behavior (Dynamic Degree, Camera 
Motion). We also report VBench-I2V's Subject/Background metrics, which measure reference-to-video consistency.

\subsection{Qualitative Experiments}

\noindent \textbf{Qualitative Results.}
Our method injects a broad range of concept categories into a generated 
world, including living entities such as cowboy and bear, objects such as cauldron and camping tent, and physically 
complex elements such as fire and glass (Fig.~\ref{fig:qual_ours}). Under image-based control (Fig.~\ref{fig:qual_ours}~(a)), the method 
preserves the visual identity of the reference image, retaining 
fine-grained details of the spawned concept throughout the rollout. See Appendix~\ref{supp:qual} for more examples.

Fig.~\ref{fig:use_case} highlights three workflows enabled by our 
method. Our approach supports spawning multiple distinct concepts at 
different points within a single rollout, with each concept persisting 
through subsequent generation. In Fig.~\ref{fig:use_case}~(a), three 
concepts are introduced sequentially into a 30-second rollout of a 
snowy forest world. A key property is illustrated here: the lamppost 
is injected only once, yet reappears after the camera turns away and 
returns to face it (Fig.~\ref{fig:use_case}~(a), final frames), 
showing that the concept propagates through temporal context and is 
naturally retrieved when the camera returns to the corresponding 
viewpoint. Our method also supports a story-driven workflow 
(Fig.~\ref{fig:use_case}~(b)), where a panorama generated from a short 
story description seeds both the reference image and the concepts that 
populate it, turning an illustrated scene into a navigable world that 
the user can traverse with concepts appearing at their authored 
locations. Beyond individual entities, our method enables broader environmental 
modifications. In Fig.~\ref{fig:use_case}~(c), the base model defaults 
to generating additional buildings consistent with the coastal village 
reference, even with diverse prompting; 
our framework instead spawns diverse landscape features such as a beach, a church, an ocean, and a road, producing distinctly different 
navigable worlds.

\begin{figure}
    \centering
    \includegraphics[width=0.95\linewidth]{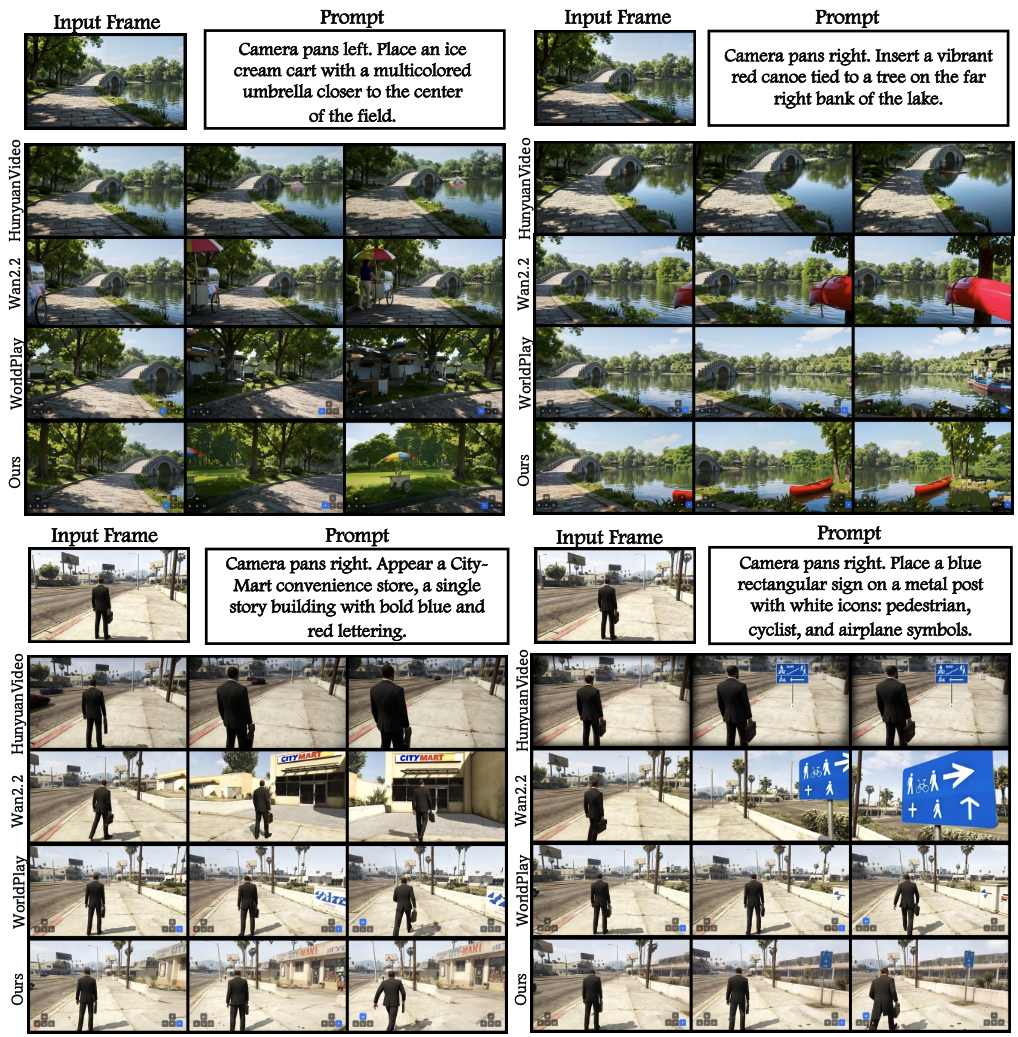}
    \caption{Qualitative comparison against baselines. Our method produces correct 
placement and action. Baselines omit the concept (HunyuanVideo, WorldPlay) 
or place it incorrectly (Wan 2.2).}
    \label{fig:qual_comp}
    \vspace{-1.5em}
\end{figure}
\noindent \textbf{Qualitative Comparison.}
Figure~\ref{fig:qual_comp} compares our method against HunyuanVideo, 
Wan~2.2, and WorldPlay on four scenes, each pairing a camera motion 
with a prompt-specified concept insertion across small props, large 
objects, buildings, and signage. HunyuanVideo omits the requested 
concept across all four scenes. Wan~2.2 introduces the concept but 
renders it inaccurately: in the canoe scene 
(Fig.~\ref{fig:qual_comp}, top-right), the canoe appears in the 
foreground rather than the far right bank, and in the sign scene 
(bottom-right), the rendered icons do not match the prompt. WorldPlay 
executes the camera trajectory faithfully but typically fails to 
render the concept (e.g., no convenience store in 
Fig.~\ref{fig:qual_comp}, bottom-left). Our method introduces the 
requested concept in every scene with correct placement, consistent 
lighting and perspective, and a preserved camera trajectory.

\begin{table*}[t]
\centering
\tiny
\caption{Evaluation across VBench metrics: comparison against baselines (top) and ablation study (bottom). Columns: 
\textbf{AQ}: Aesthetic Quality, \textbf{BC}: Background Consistency, 
\textbf{DD}: Dynamic Degree, \textbf{IQ}: Imaging Quality, \textbf{MS}: 
Motion Smoothness, \textbf{SC}: Subject Consistency, \textbf{CM}: Camera 
Motion, \textbf{I2VS}: I2V Subject, \textbf{I2VB}: I2V
Background. Best in \textbf{bold}, second-best 
\underline{underlined}.}
\label{tab:vbench_eval}
\renewcommand{\arraystretch}{1.3}
\setlength{\tabcolsep}{4pt}
\resizebox{\textwidth}{!}{%
\begin{tabular}{lccccccccccc}
\toprule
Method & AQ & BC & DD & IQ& MS & SC & CM & I2VS & I2VB & Overall\\
\midrule
\multicolumn{10}{l}{\textit{Comparison}} \\
\midrule
Hunyuan & \underline{0.640} & \underline{0.928} & 0.565 & 0.745 & \textbf{0.989} & \underline{0.893} & 0.445 & 0.964 & 0.969 & 0.749 \\
Wan 2.2 & \textbf{0.704} & \textbf{0.960} & \underline{0.670} & \textbf{0.755} & 0.981 & \textbf{0.919} & \underline{0.860} & \textbf{0.982} & \textbf{0.988} & \underline{0.853}
\\
WorldPlay   & 0.601 & 0.923 & \textbf{1.000} & 0.735 & \underline{0.985} & 0.880 & \textbf{1.000} & \underline{0.968} & 0.969 & 0.846\\
\textbf{Ours}  & \underline{0.640} & 0.924 & \textbf{1.000} & \underline{0.749} & \underline{0.985} & 0.862 & \textbf{1.000}  & 0.965  & \underline{0.981} & \textbf{0.901}\\
\midrule
\multicolumn{10}{l}{\textit{Ablation: which slots to replace?}} \\
\midrule
Slot 0      & 0.616 & 0.914 & 1.000 & 0.747 & 0.983 & 0.862 & 1.000  & 0.965 & 0.969 & 0.893 \\
Slot 1      & 0.606 & 0.920 & 1.000 & 0.735 & 0.984 & 0.875 & 1.000  & 0.967 & 0.965 & 0.892 \\
All anchors & 0.617 & 0.914 & 1.000 & 0.746 & 0.983 & 0.865 & 1.000  & 0.852 & 0.966 & 0.860  \\
Middle slot & 0.606 & 0.923 & 1.000 & 0.736 & 0.984 & 0.883 & 1.000  & 0.967 & 0.966 & 0.845 \\
Last slot & 0.603 & 0.901 & 1.000 & 0.747 & 0.982 & 0.832 & 1.000  & 0.835 & 0.813 & 0.751 \\
Slot 0 w/o pano & 0.602 & 0.924 & 1.000 & 0.749 & 0.982 & 0.878 & 1.000  & 0.968 & 0.967 & 0.894 \\
Last w/o pano & 0.596 & 0.912 & 1.000 & 0.748 & 0.982 & 0.844 & 1.000 & 0.888 & 0.896 & 0.843\\

\bottomrule
\end{tabular}%
}
\end{table*}
\subsection{Quantitative Results}
Table~\ref{tab:vbench_eval} reports VBench metrics across the four 
methods. Our method achieves the highest Overall Score and leads on 
Dynamic Degree and Camera Motion, where HunyuanVideo and Wan~2.2 
score substantially lower than our method. Subject Consistency and I2V Subject/Background reward rollouts whose 
visual content remains close to the reference or previous frame. 
Concept spawning, by construction, introduces content not present in 
the reference frame, so reductions on these dimensions are a predictable consequence. Our method nonetheless improves 
over WorldPlay on perceptual-quality dimensions, including Aesthetic Quality and Imaging Quality.

To directly measure concept fidelity, we also compute 
\emph{I2V Subject/Background} with the prepared concept image in place 
of the reference image, capturing how consistently the concept itself 
is rendered. Our method scores 0.885 on concept-conditioned I2V Subject and 
0.907 on I2V Background under this configuration; the 
baselines do not accept a concept image as input, so this metric 
applies only to our method. Overall, our method successfully 
introduces user-specified concepts into the generated world, improves 
over the base model on perceptual quality, and achieves the highest 
Overall Score.
\begin{wrapfigure}{r}{0.5\linewidth}
\vspace{5pt}
    \centering
    \includegraphics[width=\linewidth]{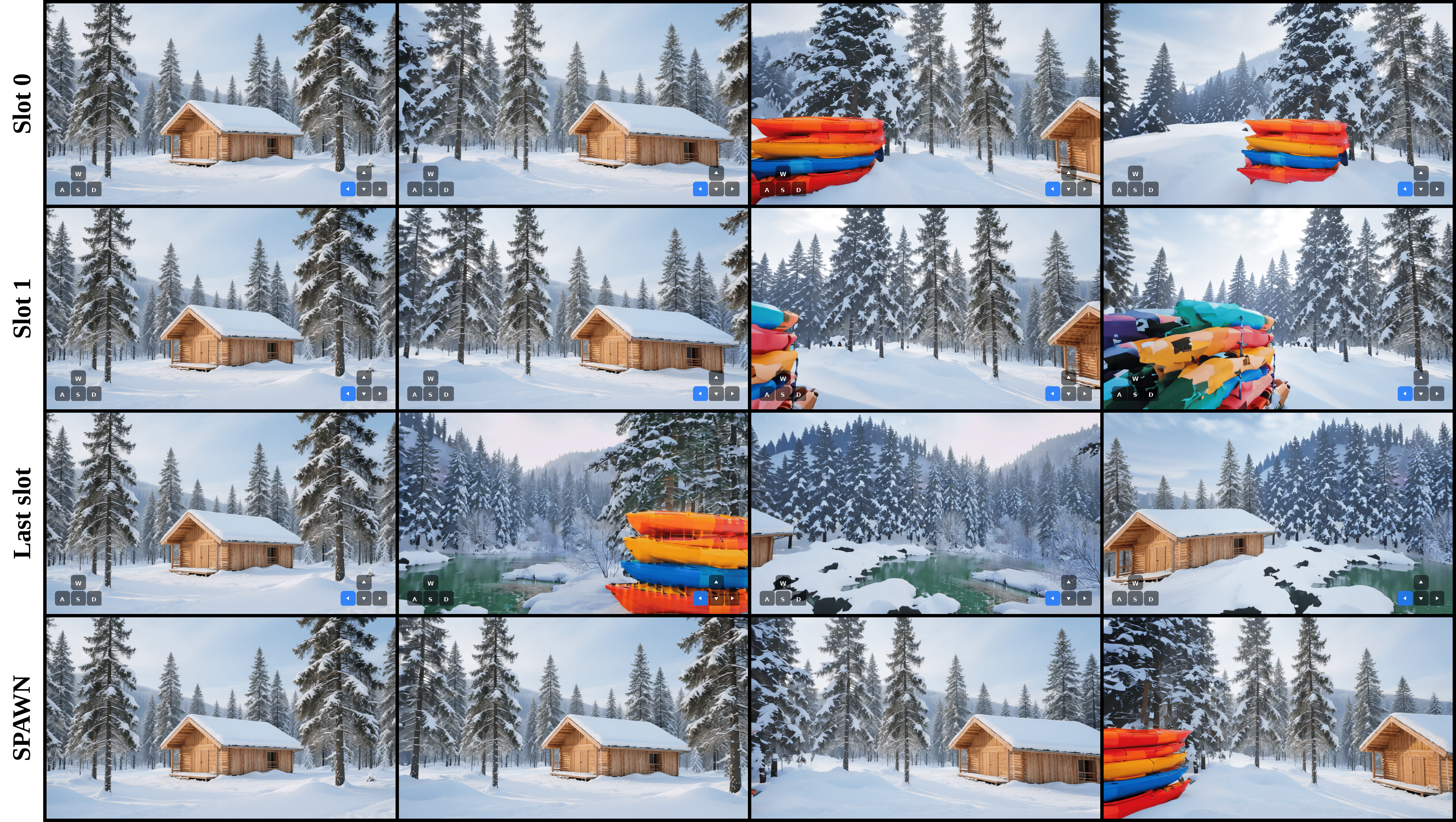}
    \caption{Slot replacement ablation.}
    \label{fig:ablation}
\end{wrapfigure}

\noindent \textbf{Ablation Study} We ablate our two-slot design against alternative slot-replacement 
strategies (Table~\ref{tab:vbench_eval}, bottom; Fig.~\ref{fig:ablation}). 
\textbf{Slot 0} spawns the concept but with weaker identity preservation 
than our two-slot variant. \textbf{Slot 1} fails to spawn the concept correctly, 
and the \textbf{last slot} causes it to appear abruptly, disrupting 
scene integration. Without panoramic grounding, the model lacks 
spatial guidance and produces 
inconsistent placement. Our two-slot variant (\textbf{Slots 0+1}) 
introduces the concept smoothly with strong identity preservation. The full ablation is provided in Appendix~\ref{supp:ablation}, including slot replacement, panoramic grounding, and the number of injection chunks settings.

\begin{wraptable}{r}{0.5\linewidth}
\centering
\tiny
\setlength{\tabcolsep}{4pt}
\caption{User study results. Rating on a 1-5 scale.}
\label{tab:user_study}
\begin{tabular}{lccc|c}
\toprule
Method & Quality $\uparrow$ & Prompt $\uparrow$ & 
Motion $\uparrow$ & Img. Fidelity $\uparrow$ \\
\midrule
Hunyuan & 3.180$\pm$1.15 & 1.554$\pm$1.06 & 1.387$\pm$0.94 & N/A \\
Wan 2.2       & 3.219$\pm$1.15 & 2.393$\pm$1.34 & 3.058$\pm$1.42 & N/A \\
WorldPlay    & 3.058$\pm$1.14 & 1.458$\pm$1.01 & 4.438$\pm$0.87 & N/A \\
\textbf{Ours} & \textbf{3.283$\pm$1.21} & \textbf{3.961$\pm$1.13} & 
\textbf{4.587$\pm$0.72} & \textbf{3.836$\pm$0.96}\\
\bottomrule
\end{tabular}
\end{wraptable}

\noindent \textbf{User Study} We conducted a user study on 20 videos with 30 raters 
to validate the results with human judgment. Raters scored each video on a 1–5 
scale across visual quality, prompt following, motion control, and concept image fidelity (baselines that do not support concept image conditioning are marked N/A).  Table~\ref{tab:user_study} shows that our method achieves the 
highest score on all criteria, with the largest 
margin on prompt following. See Appendix~\ref{supp:user} for details.

\section{Limitations}
\label{sec:limitations}

Our method operates entirely at inference time on top of a pretrained 
autoregressive world model. The available camera trajectories, 
interaction primitives, and supported action space are therefore 
inherited from the underlying model, and concept spawning does not 
extend them. For example, when the base model cannot reliably execute 
a long forward trajectory through the scene, our 
method cannot spawn concepts at locations beyond that reachable region.

\section{Conclusion}
\label{sec:conclusion}

We presented \textit{SPAWN}, a training-free method for concept spawning in autoregressive world models. \textit{SPAWN} exploits the structural anchor role of the first context-memory slot, swapping it with an external concept latent over a short windowed injection before restoring the original anchor. The injected concepts integrate into the scene with consistent lighting, scale, and perspective, while preserving visual identity and temporal coherence across extended rollouts and supporting multiple sequential injections. In doing so, we unlock a latent capacity for concept spawning that already exists in autoregressive world models built on image-to-video backbones, opening a path toward controllable concept introduction without the data and training costs of learning-based approaches.

\section*{Acknowledgements}

This work is supported by the National Science Foundation under Grant No.\ 2543524. We also gratefully acknowledge \href{https://fal.ai}{fal.ai} for providing compute support.

\bibliographystyle{splncs04}
\bibliography{main}

\clearpage
\appendix

\section*{Table of Contents}
\addcontentsline{toc}{section}{Supplementary Material Table of Contents}
\startcontents[appendix]
\printcontents[appendix]{l}{1}{\setcounter{tocdepth}{2}}

\newpage 

\section{Supplementary material}
\begin{figure}
    \centering
    \includegraphics[width=\linewidth]{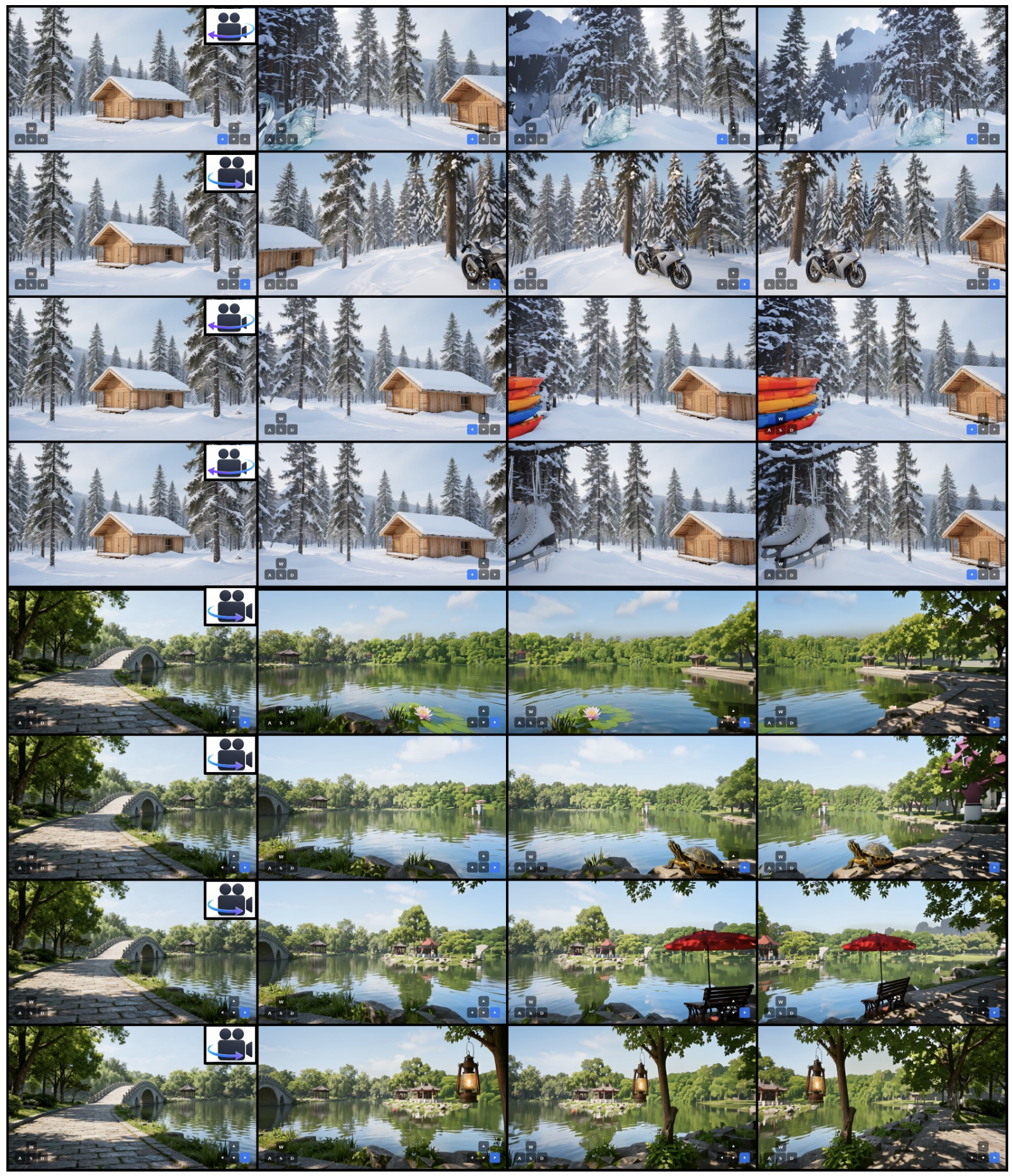}
    \caption{Additional qualitative results.}
    \label{fig:supp_qual}
\end{figure}

\subsection{Experiment Details}
\label{supp:exp}

To evaluate SPAWN at scale across 200 rollouts spanning diverse scenes and concept categories, we automate the construction of evaluation prompts and concept images. For each scene in our evaluation set, we use GPT-4o~\cite{achiam2023gpt} to generate a diverse set of concepts that are physically plausible and visually grounded in the scene context, spanning categories such as people, animals, vehicles, street fixtures, signage, and natural elements. The model is prompted to write specific edit instructions describing what to insert and where, with explicit constraints to avoid abstract or surreal elements and to ensure each prompt introduces a distinct concept.

Given the resulting prompt set, we render the corresponding concept images by editing the panoramic crop of each scene with Qwen-Image-Edit~\cite{wu2025qwenimagetechnicalreport}. This pipeline allows us to construct the full evaluation set without manual curation while maintaining concept diversity and grounding.

\subsection{Additional Qualitative Results}
\label{supp:qual}
Figure~\ref{fig:supp_qual} shows additional rollouts from SPAWN across two scenes and a range of concept types. In the snowy forest scene (rows 1--4), we spawn an ice sculpture, a motorcycle, a stack of canoes, and a pair of ice skates, each rendered with consistent lighting and snow accumulation appropriate to the environment. In the lakeside park scene (rows 5--8), we spawn a lily pad with a blooming flower, a turtle on a rock, a red umbrella with a bench, and a hanging lantern, each integrated into the scene with plausible placement and reflections on the water surface. Across both scenes, the spawned concepts persist as the camera continues its trajectory and remain coherent with the surrounding environment, demonstrating that SPAWN generalizes across concept categories from small props to environmental fixtures without any per-concept tuning.

\subsection{Ablation Study}
\label{supp:ablation}
\begin{figure}
    \centering
    \includegraphics[width=0.95\linewidth]{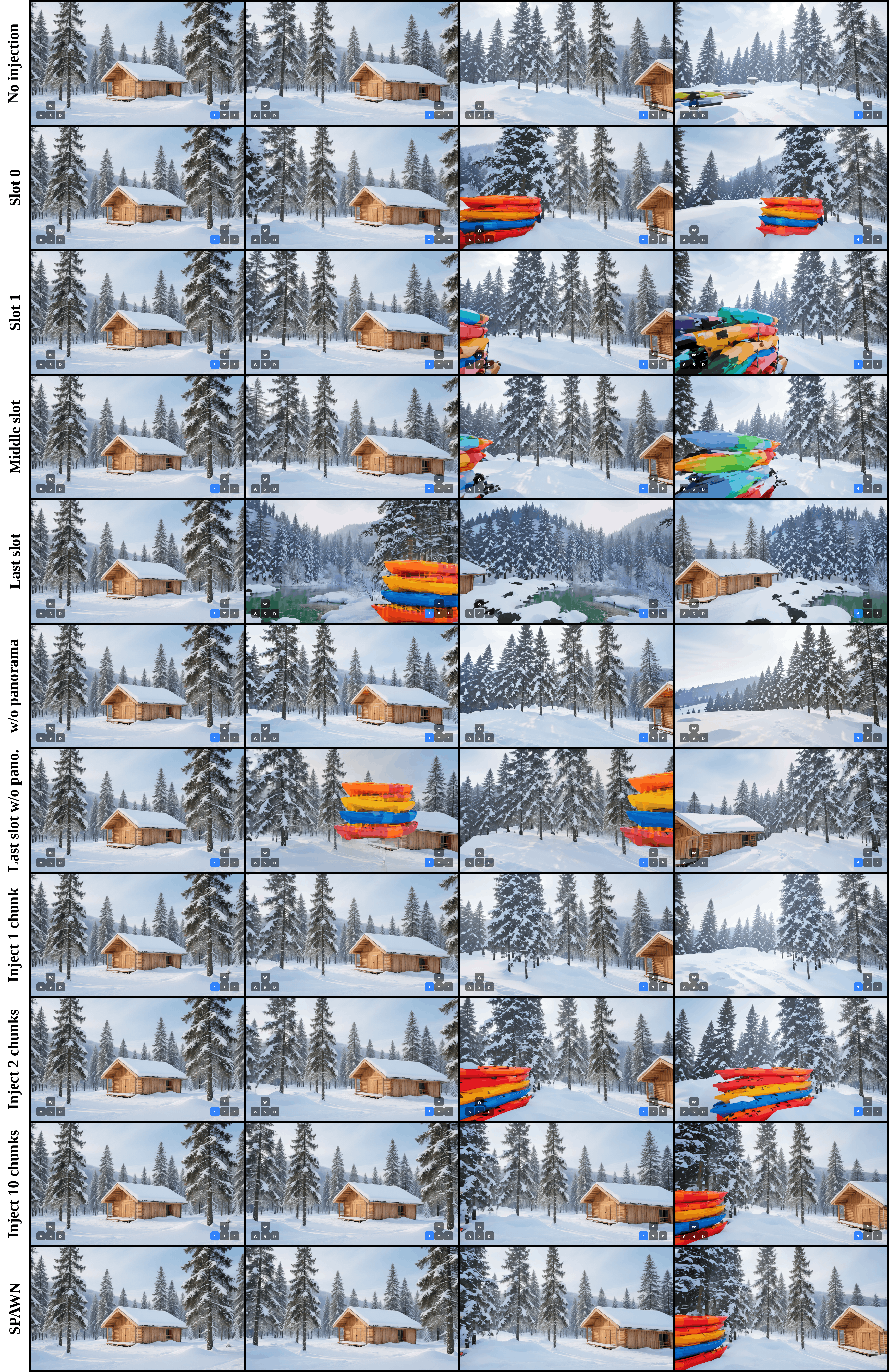}
    \caption{Qualitative ablation across slot choice, panoramic grounding, and injection window length.}
    \label{fig:ablation1}
\end{figure}

\begin{figure}
    \centering
    \includegraphics[width=0.95\linewidth]{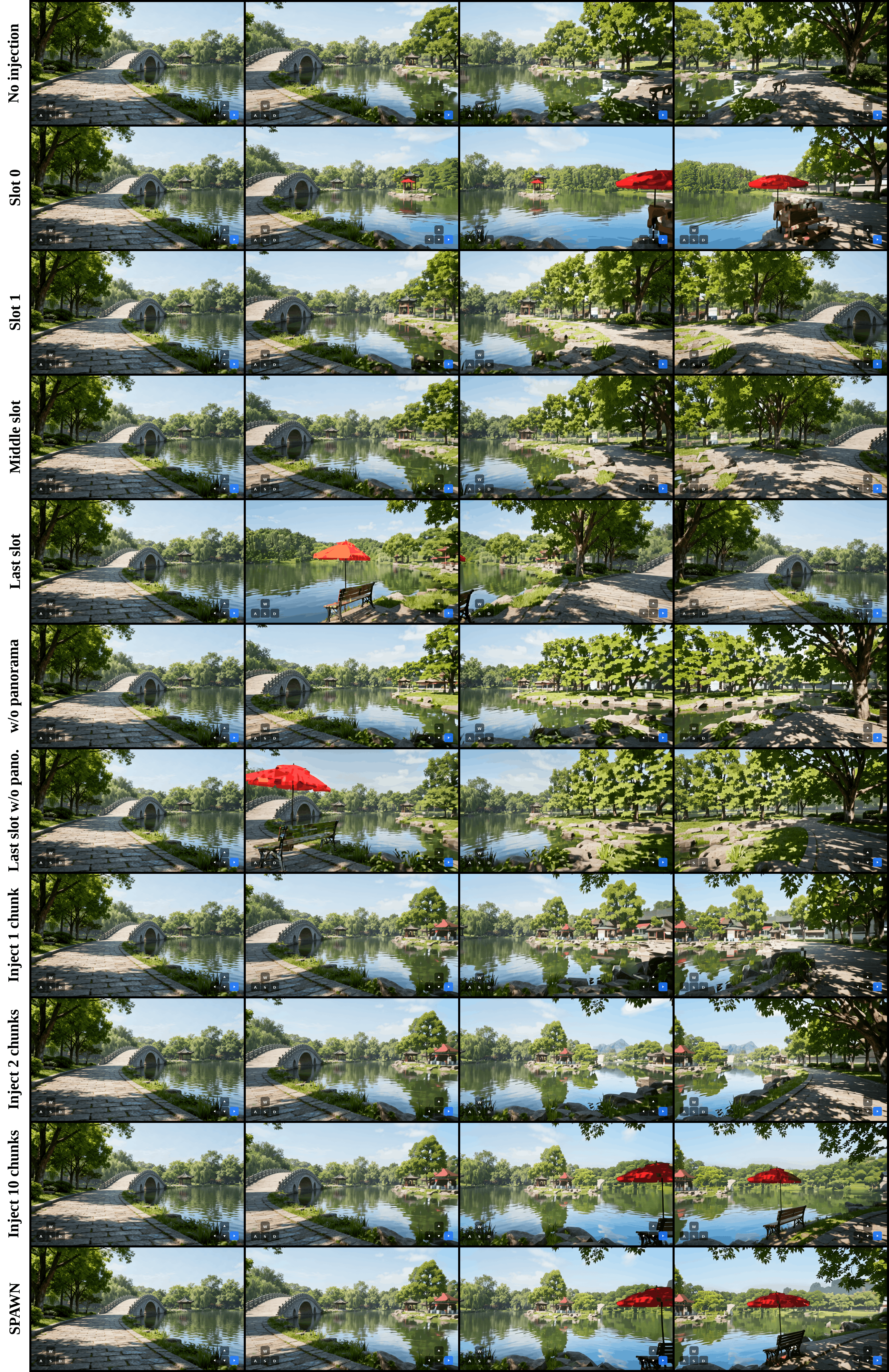}
    \caption{Qualitative ablation across slot choice, panoramic grounding, and injection window length.}
    \label{fig:ablation2}
\end{figure}

We ablate three design choices: which slots to replace, the role of panoramic grounding, and the length of the injection window (Table~\ref{tab:vbench_supp}; see Fig.~\ref{fig:ablation1} and Fig.~\ref{fig:ablation2} for qualitative results). The I2VSC and I2VBC columns directly measure concept fidelity by comparing the rollout against the concept image, and provide the clearest signal across these ablations.

\noindent\textbf{Which slots to replace.} Replacing Slot 0 alone spawns the concept but with weaker identity preservation (I2VSC 0.876, I2VBC 0.860) than our two-slot variant. Replacing Slot 1 alone or a middle slot fails to spawn the concept, reflected in the sharp drop in concept fidelity (I2VSC $\approx$0.82, I2VBC $\approx$0.80). Replacing all anchors yields the highest raw concept similarity (I2VSC 0.963), but at the cost of scene fidelity (I2V Subject collapses to 0.852) and overall quality. Replacing the last slot performs worst on every concept fidelity metric, with the concept appearing abruptly and failing to integrate. Our two-slot variant (Slots 0+1) introduces the concept smoothly while preserving identity (I2VSC 0.885, I2VBC 0.907).

\noindent\textbf{Panorama cut.} Without panoramic grounding, the concept image is the bare concept on 
a blank background, with no surrounding scene context. In this setting, concept fidelity drops sharply (I2VSC $\approx$0.65, I2VBC $\approx$0.67 for both slot configurations), since the model has no spatial context to ground the concept in the scene. Panoramic grounding provides the surrounding scene context that allows the anchor substitution to take hold.

\noindent\textbf{Number of injection chunks.} As shown in Table~\ref{tab:vbench_supp}, general video-quality and 
scene-consistency metrics remain stable across injection windows of 
1, 2, and 10 chunks. The metric that responds is concept fidelity 
(I2VSC/I2VBC), which rises from 0.793/0.801 at a single chunk and 
peaks at our 5-chunk setting (0.885/0.907) before declining slightly 
at 10 chunks (0.870/0.885). We therefore use 5 chunks: long enough to 
establish the concept faithfully, yet short enough to avoid the 
subject-consistency and concept-fidelity drop seen at 10, where the 
prolonged injection begins to crowd out scene context. This setting 
also yields the highest Overall score (0.901).

Combining these choices, Slots 0+1, 5-chunk injection, and panoramic grounding, yields our final SPAWN configuration, which introduces the concept smoothly while preserving identity and scene coherence.

\begin{table*}[t]
\centering
\tiny
\caption{Evaluation across VBench metrics. Columns: 
\textbf{AQ}: Aesthetic Quality, \textbf{BC}: Background Consistency, 
\textbf{DD}: Dynamic Degree, \textbf{IQ}: Imaging Quality, \textbf{MS}: 
Motion Smoothness, \textbf{SC}: Subject Consistency, \textbf{CM}: Camera 
Motion, \textbf{I2VS}: I2V Subject, \textbf{I2VB}: I2V
Background, \textbf{I2VSC}: I2V Subject (Concept), \textbf{I2VBC}: I2V
Background (Concept). Best in \textbf{bold}.}
\label{tab:vbench_supp}
\renewcommand{\arraystretch}{1.3}
\setlength{\tabcolsep}{4pt}
\resizebox{\textwidth}{!}{%
\begin{tabular}{lccccccccccccc}
\toprule
Method & AQ & BC & DD & IQ& MS & SC & CM & I2VS & I2VB &  I2VSC & I2VBC &Overall\\
\midrule
\multicolumn{12}{l}{\textit{Which slots to replace?}} \\
\midrule
Slot 0      & 0.616 & 0.914 & 1.000 & 0.747 & 0.983 & 0.862 & 1.000  & 0.965 & 0.969 & 0.876 & 0.860 & 0.893 \\
Slot 1      & 0.606 & 0.920 & 1.000 & 0.735 & 0.984 & 0.875 & 1.000  & 0.967 & 0.965 & 0.819 & 0.798 & 0.892 \\
All anchors & 0.617 & 0.914 & 1.000 & 0.746 & 0.983 & 0.865 & 1.000  & 0.852 & 0.966 & \textbf{0.963} & \textbf{0.966} & 0.860  \\
Middle & 0.606 & 0.923 & 1.000 & 0.736 & 0.984 & 0.883 & 1.000  & 0.967 & 0.966 & 0.815 & 0.798 & 0.845 \\
Last  & 0.603 & 0.901 & 1.000 & 0.747 & 0.982 & 0.832 & 1.000  & 0.835 & 0.813 & 0.763 & 0.719 & 0.751 \\
\midrule
\multicolumn{12}{l}{\textit{Effect of panorama cut}} \\
\midrule
Slot 0 w/o pano. & 0.602 & 0.924 & 1.000 & \textbf{0.749} & 0.982 & \textbf{0.878} & 1.000  & \textbf{0.968} & 0.967 & 0.651 & 0.666 & 0.894 \\
Last w/o pano. & 0.596 & 0.912 & 1.000 & 0.748 & 0.982 & 0.844 & 1.000 & 0.888 & 0.896 & 0.649 & 0.669 & 0.843\\
\midrule
\multicolumn{12}{l}{\textit{How many chunks to inject?}} \\
\midrule
1 chunk & 0.608 & 0.919 & 1.000 & 0.744 & \textbf{0.985} & 0.874 & 1.000 & 0.965 & 0.969 & 0.793 & 0.801 & 0.893 \\
2 chunks & 0.613 & 0.916 & 1.000 & 0.745 & \textbf{0.985} & 0.873 & 1.000 & 0.967 & 0.968 &  0.815 & 0.821 & 0.894 \\
10 chunks & 0.624 & 0.914 & 1.000 & 0.748 & 0.984 & 0.856 & 1.000 & 0.963 & 0.966 & 0.870 & 0.885 & 0.891\\
\midrule
\multicolumn{12}{l}{\textit{SPAWN: slots 0+1, 5 chunks, with panorama}} \\
\midrule
SPAWN & \textbf{0.640} & \textbf{0.924} & \textbf{1.000} & \textbf{0.749} & \textbf{0.985} & 0.862 & \textbf{1.000}  & 0.965  & \textbf{0.981} & 0.885 & 0.907 & \textbf{0.901}\\
\bottomrule
\end{tabular}%
}
\end{table*}

\subsection{User Study Details}
\label{supp:user}
\begin{figure}
    \centering
    \includegraphics[width=0.7\linewidth]{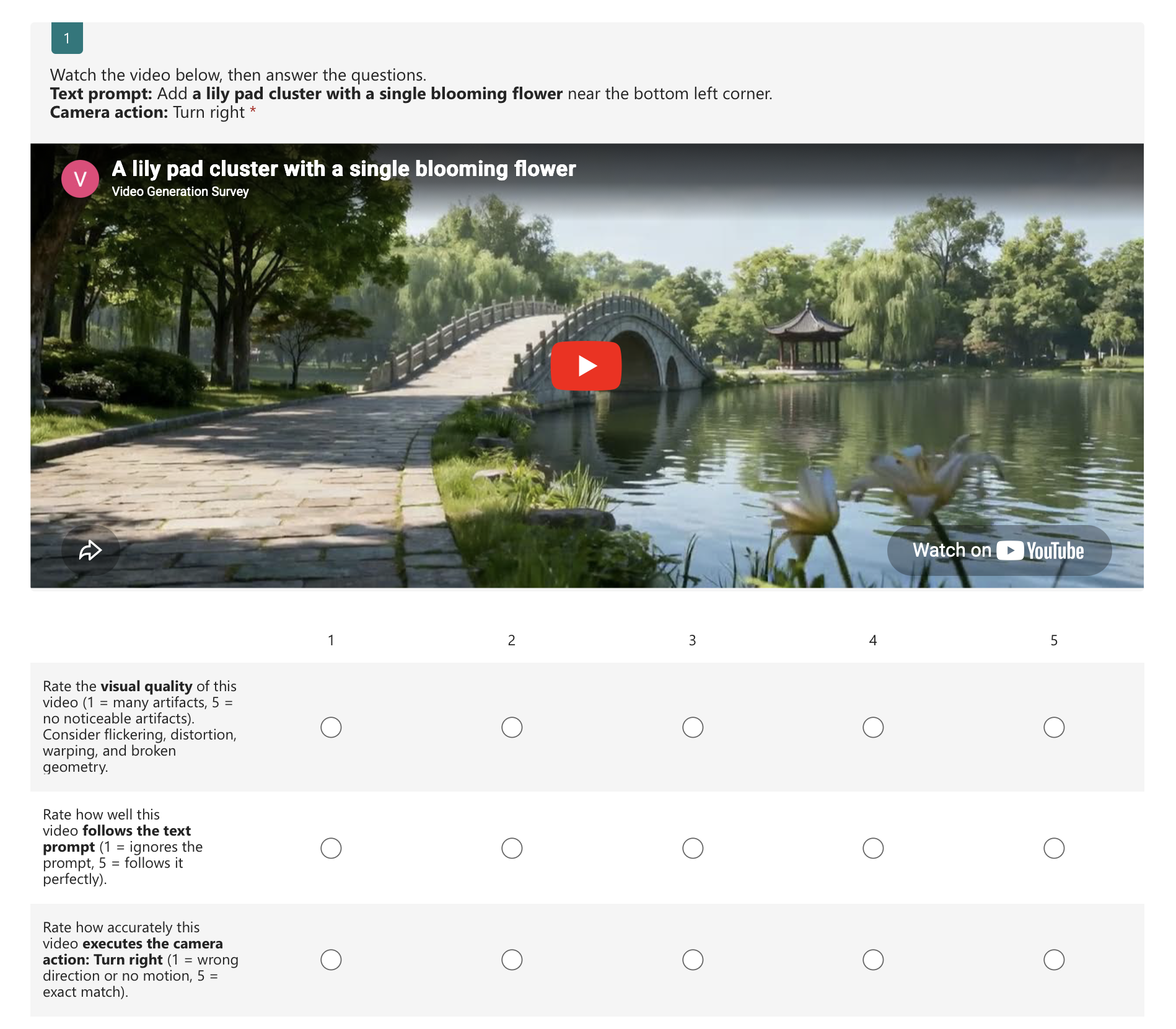}
    \caption{User study interface: raters scored each video on visual quality, prompt following, and motion control.}
    \label{fig:user1}
\end{figure}
\noindent\textbf{Study design.} We recruited 30 raters through Prolific, compensated in line with Prolific's fair pay guidelines. Each rater evaluated 20 generated videos sampled from our evaluation set, covering a range of scenes, concepts, and visual styles. For each video, raters were shown the text prompt describing the requested concept and the camera action, and where applicable, the concept image used to drive SPAWN. Videos from all four methods (HunyuanVideo, Wan~2.2, WorldPlay, and SPAWN) were included and presented in randomized order to mitigate ordering effects.

\noindent\textbf{Rating dimensions.} Raters scored each video on a 1--5 scale across four dimensions:
\begin{itemize}
    \item \textbf{Visual quality.} \emph{``Rate the visual quality of this video (1 = many artifacts, 5 = no noticeable artifacts). Consider flickering, distortion, warping, and broken geometry.''}
    \item \textbf{Prompt following.} \emph{``Rate how well this video follows the text prompt (1 = ignores the prompt, 5 = follows it perfectly).''}
    \item \textbf{Camera action.} \emph{``Rate how accurately this video executes the camera action (1 = wrong direction or no motion, 5 = exact match).''} The specific camera action (e.g., \emph{turn right}, \emph{pan left}) was shown alongside the question.
    \item \textbf{Concept image fidelity.} For SPAWN only, raters were shown the concept image alongside the video and asked to rate how faithfully the concept appears in the rollout, on a 5-point scale from \emph{Very poor} to \emph{Excellent}. Baselines do not accept a concept image as input and were marked N/A on this dimension.
\end{itemize}

\begin{figure}
    \centering
    \includegraphics[width=0.7\linewidth]{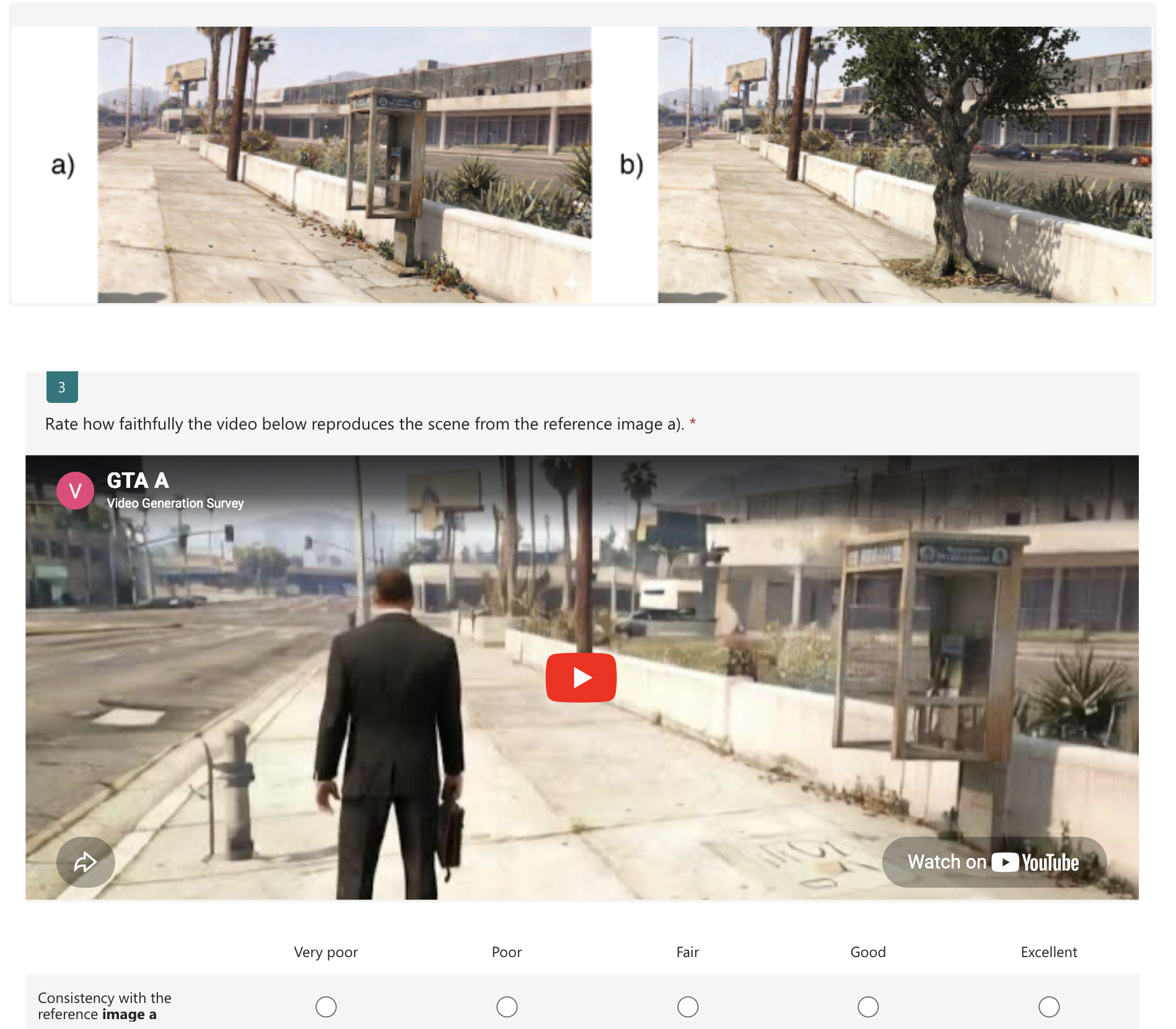}
    \caption{User study interface: raters scored each video on concept-image fidelity.}
    \label{fig:user2}
\end{figure}

\noindent\textbf{Interface.} Each video was embedded inline above its rating questions, with the text prompt and camera action presented immediately above the video. Concept-fidelity questions additionally displayed the reference image and concept image side by side, labelled \emph{a)} and \emph{b)}, with the rollout below. Screenshots of the rating interface are shown in Figures~\ref{fig:user1} and ~\ref{fig:user2}.

\noindent\textbf{Results.} Across all four dimensions, SPAWN scores highest, with the largest margin on prompt following (3.961 vs.\ 2.393 for the next-best method). The visual-quality scores are tightly clustered across methods (3.06--3.28), indicating that none of the rollouts exhibit catastrophic artifacts and that the differences between methods are concentrated in controllability rather than image quality. Motion-control scores reflect the gap between general I2V backbones and autoregressive world models: HunyuanVideo and Wan~2.2, which encode camera motion through the text prompt rather than an action interface, score substantially lower (1.39 and 3.06) than WorldPlay and SPAWN (4.44 and 4.59), consistent with the camera-motion findings in our automatic evaluation. Concept-image fidelity, applicable only to SPAWN, averages 3.84, confirming that the concept rendered in the rollout is recognizably the one provided as input.

\end{document}